# Cascade Convolutional Neural Network for Image Super-Resolution

Jianwei Zhang, Zhenxing Wang, Yuhui Zheng, and Guoqing Zhang*

School of mathematics and statistics, Nanjing University of Information Science and Technology, Nanjing, 210044, China.

**Abstract**：With the development of the super-resolution convolutional neural network (SRCNN), deep learning technique has been applied to the field of image super-resolution communities. Many researchers focus on optimizing and improving the structure of SRCNN, which achieved well performance in speed and restoration quality for image super-resolution. However, most of these previous approaches only consider a specific scale images during the training process, while ignoring the relationship between different scale images. Motivated by this concern, in this paper, we propose a cascade convolution neural network (CSRCNN) for image super-resolution, which includes three cascaded Fast SRCNN and each Fast SRCNN can process a specific scale image. Thus, different scale images can be trained at the same time and the learned network can make full use of the information reside in different scale images. Extensive experiments have shown that our method performs better than the state-of-the-art methods.

## 1 Introduction

Single image super-resolution (SR) is an important issue in the field of computer vision research. Given a low resolution (LR) image, the purpose of single image super-resolution is to recovering a high resolution (HR) image corresponding to the LR image. Currently, single image super-resolution based methods are generally divided into two categories: sparse coding-based methods (Yang and Yang., 2013; Farhadifard et al., 2014; Lin-Yuan et al., 2016; Ahmed et al., 2018; Yang et al., 2019) and deep learning-based methods (Cui et al., 2014; Choi et al., 2018; Dong et al., 2014, 2016).

Most sparse coding-based methods assume that each pair of patches in LR and HR images have similar coding coefficients in the patch space (Yang et al., 2010; Farhadifard et al., 2014; Ahmed et al., 2016, 2017). Therefore, the HR image patches can be represented by the sparse coding LR image patches. Specifically, learning a pair of dictionaries $D_L$ and $D_H$ for LR and HR image patches respectively (Yang et al., 2008, 2012). The final reconstructed HR image is obtained through the learned dictionary $D_H$ and coding $\alpha$ obtained by the learned dictionary $D_L$ (Zhang et al.,2017, 2019; Zheng et al., 2019). At present, existing sparse coding-based methods mainly focus on how to learn better dictionaries and optimize dictionaries (Zhao et al., 2018; Ahmed et al., 2018; Yang et al., 2019;). For example, Zhao et al (Zhao et al., 2018;) design a transfer robust sparse coding based on graph for image representation and Ahmed et al (Ahmed et al., 2018) propose a new multiple dictionary learning strategy, which achieved well performance in speed and restoration quality for image super-resolution.

Deep learning-based methods aims to directly learn an end-to-end mapping from LR images to HR images (Cui et al., 2014; Choi et al., 2018; Dong et al., 2014, 2016). Dong et al (2014) first applied convolutional neural network to the single image super-resolution, proposed a super-resolution convolutional neural network (SRCNN) and verified its better restoration performance. However, SRCNN only learns the mapping from bicubic interpolated LR images (not the original LR images) to HR images, the calculation cost of the network will be increased quadratically (Dong et al., 2016) with the size of HR image increases. In addition, SRCNN adopts a 5*5 convolution kernel to learn nonlinear mapping which obviously limited the learning ability of network in such a setting. To solve the above problems, Dong and Kim et al (Dong et al., 2014, 2016; Kim et al., 2016; Lai et al. 2017; Lim et al., 2017) proposed lots of approaches such as the large super-resolution convolutional neural network (SRCNN-ex), fast super-resolution convolutional neural networks (FSRCNN), very deep super-resolution networks (VDSR), Laplacian pyramid super-resolution network (LaPSNR) and enhanced deep super-resolution network (EDSR) to improve the network structure of SRCNN. In addition, Christian et al (Ledig et al., 2017) exploited generative adversarial networks (GAN) (Goodfellow et al., 2014) to handle out image super-resolution problem and proposed super-resolution generative adversarial network (SRGAN). GAN is different from convolutional neural networks (CNN), the main difference is that GAN aims to generate more realistic or HR images that are more consistent with human eye, while the aim of CNN is to faithfully restore the high-frequency information of images. Lots of improved networks based on GAN have been proposed in recent years (Wang et al., 2018; Zhang et al., 2019). These networks have solved the above problems to some extent and achieved well restoration performance.

However, all of the models mentioned above only trained the images in a specific scale and do not consider the relationships between different image scales. For example, if we set the scale factor is 4 during the training process, FSRCNN only uses the image information contained in scale of 4, and does not make full use of the complementary information resided in different scales. if we can make full use of the information of different scale images in the training process, the quality of recovered images will be further improved.

Based on the above ideas, we design a cascaded convolutional neural networks (CSRCNN). It consists of three cascaded FSRCNN, where each FSRCNN can process a specific scale image. For each FSRCNN, we set its scaling factor is 2 which can double enlarge the size of input image. Suppose that all person images have the same weight-to-height ratio. $W$ and $H$ respectively denote the weight and height of the original HR image. $W/8 * H/8, W/4 * H/4$ and $W/2 * H/2$ are respectively image shapes input to three subnetworks. Since images with different scale are trained together, the learned network can make use of image information of different scales. In addition, we use L1 loss function to make the reconstructed image clearer in texture and edge in the training process. Finally, we evaluated the impact of the number of cascaded FSRCNN on our network performance in 4.3. Experiments show that the proposed cascade convolutional neural network can achieve better performance.

## 2 Related Work

### 2.1 Deep Learning for Image Super-Resolution

The purpose of image super-resolution is to reconstruct high resolution image from a given low resolution image one. Dong et al (2014) first applied convolutional neural network to the field of image super-resolution and proposed a super-resolution convolutional neural network (SRCNN) which achieved better performance for image restoration. Recently, deep learning technique has been widely applied to the image super-resolution communities and most of works focus on optimizing and improving the structure of SRCNN. Various network structures have been developed, such as deep network with residual learning (Kim et al., 2016), Laplace pyramid structure (Lai et al., 2017), residual block (Zhang et al., 2018), and residual dense network (Choi et al., 2018), which obtained superior performances. Besides supervised learning, unsupervised learning (Peng., 2019; Timofte et al., 2013) and reinforcement learning (Yun et al., 2017; Timofte et al., 2014) are also introduced to solve the problem of image super-resolution in recent years. Specifically, the literature (Wang et al., 2019) has a systematic description of image super resolution.

**2.2 Fast Super-Resolution Convolutional Neural Networks**

Fast super-resolution convolutional neural networks (FSRCNN) was proposed by Dong et al (2016), aiming to accelerate the previously proposed SRCNN. Compared with SRCNN, FSRCNN can directly learn the mapping from the original LR image to HR image by introducing the deconvolution layer at the end of the network. The calculation cost of the whole network has been reduced to a certain extent by the above methods. In addition, FSRCNN adds a shrinking and an expanding layer at the beginning and the end of the mapping layer respectively, to enhance the presentation capability of non-linear mapping (He et al., 2016). In the nonlinear layer, FSRCNN adopted a smaller filter (size=3*3). The whole network is designed as a compact hourglass-type CNN structure, which includes five parts: feature extraction, shrinking, non-linear mapping, expanding, deconvolution. The shrinking layer uses a filter of size 1*1 to reduce the LR feature dimension from $d$ to $s$, where $d$ is the feature dimension of LR image after feature extraction and $s$ is the number of filters ($s \leq d$). The expanding layer uses $d$ filters with a size of 1*1 to maintain consistency with the shrinking layer, which is the inverse operation of the shrinking layer. To avoid the "dead features" (Zeiler and Fergus., 2014) caused by zero gradients in $ReLU$, the author uses $PReLU$ (He et al., 2015) as the activation function after each convolution layer. Thus, a complete FSRCNN can be represented as:

$$Conv(5,d,1) - PReLU - Conv(1,s,d) - PReLU - Conv(3,s,s) - PReLU - mConv(1,d,s)$$
$$-PReLU - DeConv(9,1,d). \qquad (1)$$

# 3 Cascade Convolutional Neural Network

## 3.1 Network Structure

We propose a cascade convolutional neural network (CSRCNN) framework in this section. Fig. 1 shows the network structure of CSRCNN. which consists of three cascaded FSRCNN (Dong et al., 2016), and each FSRCNN can process a specific scale image. In each sub FSRCNN, we set its scaling factor to 2 which can double enlarge the size of input image. Suppose that all person images have the same weight-to-height ratio. $W$ and $H$ respectively denote the weight and height of the original HR image. $W/8 * H/8, W/4 * H/4$ and $W/2 * H/2$ are respectively image shapes input to three subnetworks. Here, we use $I^k$ to represent the input image of each sub FSRCNN, where $k = 0,1,2$ represents the ID of each sub FSRCNN and the scale index of the input image. We set $r_k$ to represent the scale ratio of LR image to HR image, the size of input image $I^k$ is described as $r_k W * r_k H$. The scale ratios are respectively r0 = 1/8, r1 = 1/4, r2 = 1/2, and r3 = 1. $I^{k+1} = F_k(I^k)$ is the output image of each FSRCNN that is the restoration HR image by each sub FSRCNN-k. For each cascaded FSRCNN, the output image is twice the size of input image. We use $\tilde{I}^{k+1}$ represents the real

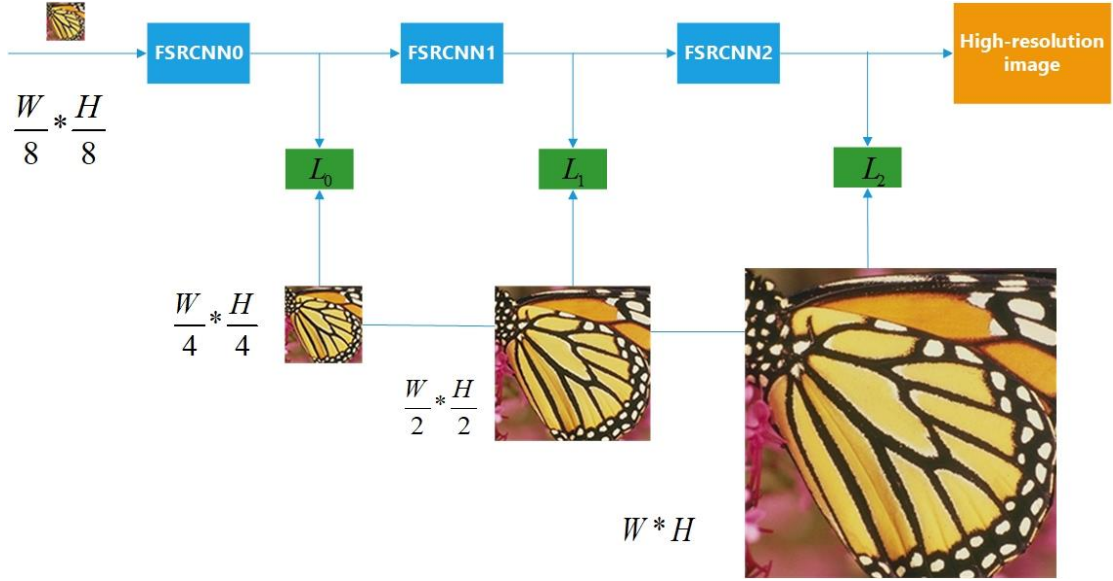

**Fig. 1**. The network structure of CSRCNN. The network is composed of three cascade FSRCNN, in which the upscaling factor of each FSRCNN is 2. The size of the input of the network is a $W/8 * H/8$ LR image, which will be successively entered into three sub-FSRCNN. Each sub-FSRCNN can double enlarge the size of input image. For each sub network, the output image of the network will form a sub loss function with the corresponding real HR image, and the sum of the sub loss functions of these sub networks will eventually form the loss function of the whole network.

HR image, which has the same size with the super-solved image $F_k(I^k)$. Our goal is to learn super resolved functions that estimate the reconstructed HR images $I^{k+1}$ and real HR images $\tilde{I}^{k+1}$

### 3.2 Loss Function

Our network is composed of three cascaded FSRCNN. For each subnetwork, on the one hand, the output image of network will enter the next cascaded subnetwork for training. On the other hand, it will form a sub loss function with the corresponding real HR image. The loss function of the whole network is composed of three sub loss functions. $L_0, L_1, L_2$ used in our experiments, represent the loss function of three subnetwork respectively. The loss function of the whole network is represented by

$$L_{loss} = L_0 + L_1 + L_2. \qquad (2)$$

For each subnetwork, the loss function is calculated as follows:

$$L_k = \frac{1}{r_{k+1}^2 WH} \sum_{x=1}^{r_{k+1}W} \sum_{y=1}^{r_{k+1}H} \left\| \hat{I}_{x,y}^{k+1} - F(I^k)_{x,y} \right\|_1. \qquad (3)$$

In the selection of loss function of each sub-network, we use L1 loss function to replace MSE loss function. The main reason is that L1 loss function make the reconstructed image clearer in texture and edge in the training process, while MSE loss function will lose high-frequency information in the reconstruction process, such as texture and edge (Yi and Huang., 2015), which makes the final reconstructed image have poor perception performance.

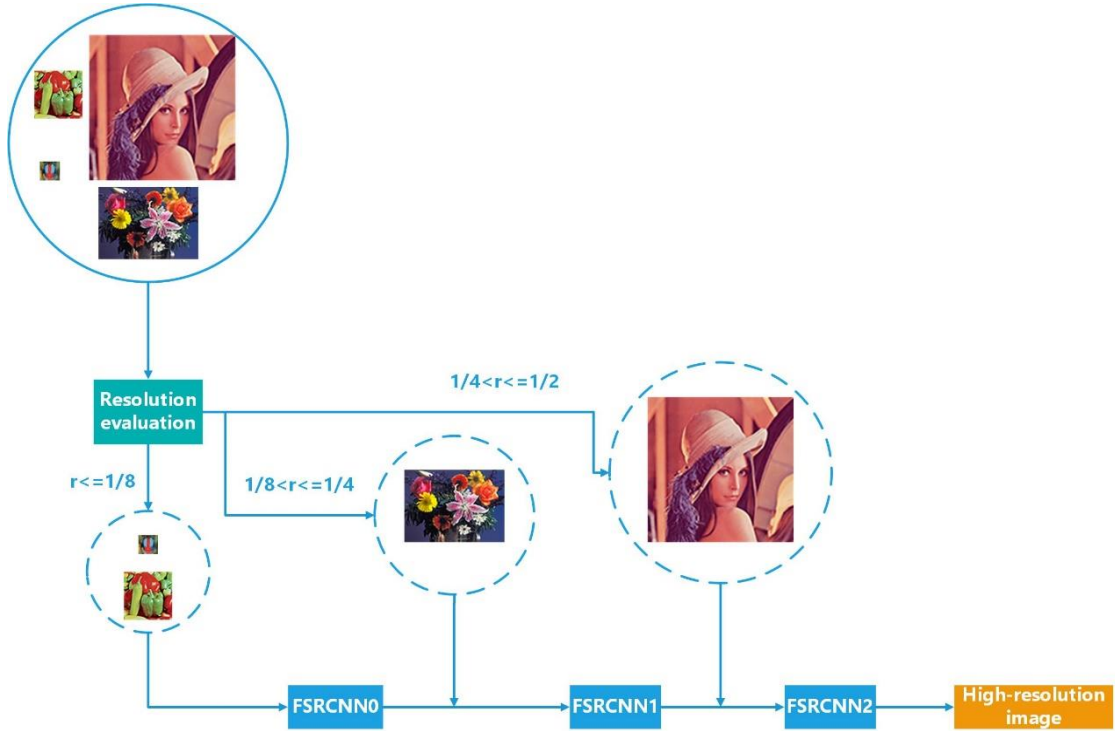

**Fig. 2.** LR images are assigned to different stages of the cascaded FSRCNN according to their resolution scale ratio. After the CSRCNN, all LR images are enlarged to the uniform HR.

### 3.3 Assessment Process

During the evaluation process, we will assign LR image to different stages of the cascaded network according to the scale ratio of LR images to HR images. When a LR image is assigned to the FSRCNN-k, it will be resized to the input shape of corresponding network $r_k W * r_k H$. For example, for a given test image with upscaling factor 3×, we will resize the shape of the image to $W/2 * H/2$. The resized image will be arranged into FSRCNN2. Here, each subnetwork can double enlarge the size of input image. At the end of the evaluation, all images are enlarged to the uniform HR image (Pang et al., 2017). Fig.2 shows the entire evaluation process for the network.

### 3.4 The Differences between FSRCNN

Our network contains three cascade FSRCNN where each FSRCNN can process a specific scale image. The loss function of each FSRCNN is L1-Loss function, while the loss function of FSRCNN (Dong et al., 2016) is MSE Loss function. In addition, FSRCNN only focus on a specific scale images in the training process, while our network can train different scales image at the same time due to the use of cascade structure. During the test stage, if we want to reconstruct two different scales images (such as 2×, 4×), we need to train two networks separately when we adopt FSRCNN, while our methods just only need to train once to get the high resolution images with scales of 2×, 4× and 8×.

## 4 Experiments

### 4.1 Dataset：

**Training and test dataset**: Following SRCNN and FSRCNN, we combined 91 images dataset and the general-100 dataset to train our network. In particular, general-100 dataset were proposed by Dong et al (2016), it contains 100 bmp-format images with ranges from 710×704 (large) to 131 × 112 (small) in size. These images are all of favorable quality with clear edges and less smooth regions. In addition, we adopted the data augmentation method (Wang, Z., 2015) to make full use of the dataset which includes scaling and rotating images. For test dataset, we use the Set5 (Bevilacqua et al., 2012), Set14 (Zeyde et al., 2010) and BSD200 dataset (Martin et al., 2001). All test images will be cropped according to the model structure which make the size of the model output image an integer.

**Training samples**: Due to our network use cascade structure, different scale images can be trained at the same time. To form LR/HR sub-image pairs, we first sample the original training images by different scale factor $n$, then crop LR images obtained in sampling process and the ground truth images.

### 4.2 Training Details

**Training strategy:** For deep model, the model performance will improve with the increase of data, so we combined 91 images dataset and the general-100 dataset (Dong et al., 2016) in training processing. First, we use the 91-image dataset to train a network from scratch. Then, we add the General-100 dataset to network for fine-tuning when the training is saturated. With this strategy, the training converges much earlier than training with the two datasets from the beginning.

**The choice of learning rate:** The performance of the network will be affected by the choice of different learning rates. Choosing a good learning rate is important for the performance of network. For FSRCNN, the learning rate of the convolution layers is set to be $10^{-3}$, the learning rate of the deconvolution layer is set to be $10^{-4}$, the learning rate of all the layers is reduced by half during fine - tuning. Obviously, the learning rate of convolution layer and deconvolutional layer is static in each iteration, while we adopt a dynamic method to update the learning rate of convolution layer and deconvolutional layer in our network which make our network produce better learning rate according to different iterations. We take $a_0 = 10^{-3}, 10^{-4}$ as the initial learning rate of the convolution layer and the deconvolution layer, and set the total number of iterations of the network as $n$. When the number of iterations is $m$, the learning rate of the network convolution layer and the deconvolution layer is represented as:

$$a_m = a_0 * 0.1^{\rho(\frac{m}{0.8*n})} \tag{4}$$

Here, $\rho\left(\frac{m}{0.8*n}\right)$ represents the integer part of the final result.

**The choice of networks initialization:** Network initialization to the training of the network has a great influence, a good initialization method can largely reduce the training time of the network. Due to our network selected *PReLU* (He et al., 2015) as the activation function, we chose the MSRA initialization (He et al., 2015) which make each layer neuron input/output variance is consistent. This is a mean of zero and variance of $\frac{n}{2}$ gaussian distribution, which makes our networks to converge faster.

**Parameter setting:** Our cascade convolutional neural network consists of three subnetworks. For each sub-network, there are three sensitive variables governing the network performance which includes the LR feature dimension $d$, the number of shrinking filters $s$, and the mapping the depth $m$. According to relevant experimental conclusions of FSRCNN, we set $d = 56, s = 12$ and $m = 4$ for each subnetwork.

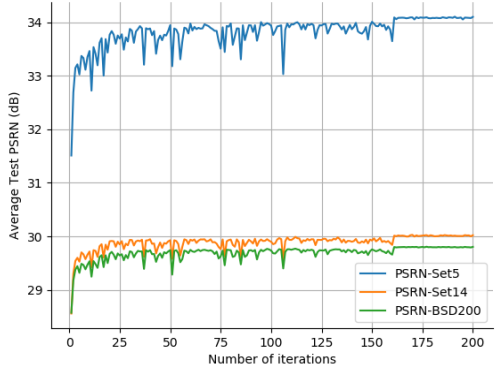

**Fig. 3.** visual comparisons on Set5, Set14 and BSD200 with upscaling factor 3.

Table 1: The influence of the number of cascaded FSRCNN with upscaling factor 4, where the test data set is Set5, Set14 and BSD200.

| Test-dataset | Upscaling factor | Net1 PSNR | Net2 PSNR | Net3 PSNR |
|---|---|---|---|---|
| Set-5 | ×4 | 30.55 | 30.82 | **31.01** |
| Set-14 | ×4 | 27.50 | 28.02 | **28.47** |
| BSD200 | ×4 | 26.92 | 27.42 | **27.68** |

### 4.3 Evaluation for the number of cascaded FSRCNN

In this section, we compare the impact of the number of cascaded FSRCNN on network performance. The network composed of one FSRCNN, two cascaded FSRCNN and three cascaded FSRCNN are represented as Net1, Net2 and Net3 respectively. For Net2, the loss function of the whole network is composed of L1 and L2 in 3.2 summary. We select PSNR as evaluation indicators. Table1 shows the results visually on Set5 (Bevilacqua et al., 2012), Set14 (Zeyde et al., 2010), BSD200 dataset for 4×. From table1, we observed that the PSNR value obtained by use Net3 is higher than that obtained by use Net2 and Net1 and the PSNR value obtained by use Net2 is higher than that obtained by use Net1 , which verified that use different scale image information in the training process can improve the performance of the network.

### 4.4 Comparison with State-of-the-Arts

In order to verify the well reconstruction performance of our proposed CSRCNN. We compare the proposed CSRCNN with 3 state-of-the-art SR algorithms: Bicubic, SRCNN (Dong et al., 2014), FSRCNN (Dong et al., 2016). We carry out extensive experiments on three datasets: Set5 (Bevilacqua et al., 2012), Set14 (Zeyde et al., 2010), BSD200 dataset (Martin et al., 2001), and adopted two image quality indicators to evaluate the SR images which includes PSNR and SSIM (Wang et al., 2004). Table2 shows quantitative comparisons for 2×, 3× and 4× SR. From Table2, we can observe that compared with the above methods, the proposed CSRCNN achieves the best results in three evaluation datasets for 2×, 3× and 4× SR. Fig.3 show the convergence curves of CSRCNN on these three evaluation datasets with upscaling factor 3. Among them, the PSNR value of the proposed CSRCNN is 0.51db, 1.04db and 0.46db higher than that of FSRNCNN in Set5 datasets for 2×, 3× and 4× SR. We observed that the PSNR value of CSRCNN is 1.04 higher than that of FSRCNN for 3× SR, one possible reason is that we preprocessed the LR image, we use bicubic interpolation when we threw the LR image into the network, such that the size of the interpolated image was half that of the real HR image.

Consider our network is made up of three cascade FSRCNN where each FSRCNN's upscaling factor is 2, we also made a comparison for multiple scaling factors 8 for FSRCNN. This comparison is shown Table3. Obviously, our network produces higher PSNR and SSIM values than FSRCNN due to the cascade design.

Table 2: Quantitative evaluation of state-of-the-art SR algorithms average PSNR/SSIM for scale factors 2×, 3× and 4×.

| Test-dataset | Upscaling factor | Bicubic PSNR/SSIM | SRCNN PSNR/SSIM | FSRCNN PSNR/SSIM | CSRCNN (our) PSNR/SSIM |
|---|---|---|---|---|---|
| Set-5 | ×2 | 33.66/0.9299 | 36.33/0.9521 | 36.94/0.9558 | **37.45/0.9570** |
| Set-14 | ×2 | 30.23/0.8677 | 32.15/0.9039 | 32.54/0.9088 | **34.34/0.9240** |
| BSD200 | ×2 | 29.70/0.8625 | 31.34/0.9287 | 31.73/0.9074 | **32.92/0.9122** |
| Set-5 | ×3 | 30.39/0.8682 | 32.45/0.9033 | 33.06/0.9140 | **34.10/0.9233** |
| Set-14 | ×3 | 27.54/0.7736 | 29.01/0.8145 | 29.37/0.8242 | **30.02/0.8346** |
| BSD200 | ×3 | 27.26/0.7638 | 28.27/0.8038 | 28.55/0.8137 | **29.78/0.8302** |
| Set-5 | ×4 | 28.42/0.8104 | 30.15/0.8530 | 30.55/0.8657 | **31.01/0.8702** |
| Set-14 | ×4 | 26.00/0.7019 | 27.21/0.7413 | 27.50/0.7535 | **28.47/0.7720** |
| BSD200 | ×4 | 25.97/0.6949 | 26.72/0.7291 | 26.92/0.7398 | **27.68/0.7552** |

Table 3: Quantitative evaluation of state-of-the-art SR algorithms average PSNR/SSIM for scale factors 8×.

| Test-dataset | Upscaling factor | Bicubic PSNR/SSIM | SRCNN PSNR/SSIM | FSRCNN PSNR/SSIM | CSRCNN (our) PSNR/SSIM |
|---|---|---|---|---|---|
| Set-5 | ×8 | 24.39/0.657 | 25.33/0.689 | 25.41/0.682 | **25.74/0.715** |
| Set-14 | ×8 | 23.19/0.568 | 23.85/0.593 | 23.93/0.592 | **24.30/0.614** |
| BSD200 | ×8 | 23.67/0.547 | 24.13/0.565 | 24.21/0.567 | **24.50/0.581** |

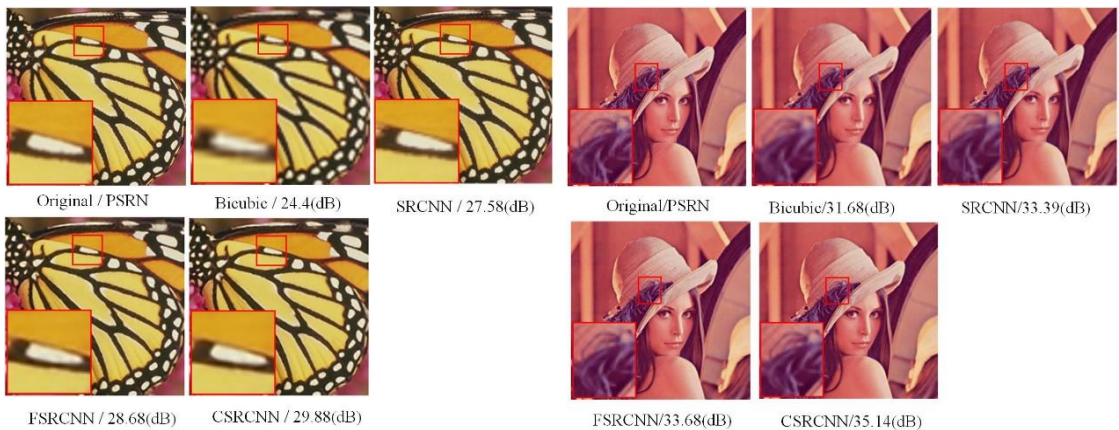

**Fig.4.** visual comparisons on Set5 and Set14 datasets with upscaling factor 3.

In Fig.4, we show visual comparisons on Set5 and Set14 datasets with upscaling factor 3. As you can see, our method accurately reconstructs the details of the image, such as texture and edges.

## 5 Conclusion

In this paper, we proposed a more efficient network for mining image information in different scales, named Cascade convolutional neural network. We used a cascaded network structure to make full use of the image information reside in different scale of images. Furthermore, we adopted L1 loss function as the loss function of each subnetwork in the training process which make the reconstructed image clearer in texture and edge. Finally, we also explore the effects of the number of cascaded FSRCNN for the performance of our network. Extensive experiments show that the proposed method achieves satisfactory SR performance.

## Acknowledgment

This research is supported in part by the National Natural Science Foundation of China under Grant 61806099, in part by the Natural Science Foundation of Jiangsu Province of China under Grant BK20180790, in part by the Natural Science Research of Jiangsu Higher Education Institutions of China under Grant 8KJB520033, in part by Startup Foundation for Introducing Talent of Nanjing University of Information Science and Technology under Grant 2243141701077.

## References

Ahmed, J., Shah, M. A., 2016. Single image super-resolution by directionally structured coupled dictionary learning. IEEE Trans. EURASIP Journal on Image and Video Processing. 36, 1-12.

Ahmed, J., Waqas, M., Ali, S., et al., 2017. Coupled dictionary learning in wavelet domain for Single-Image Super-Resolution. IEEE Trans. Signal, Image and Video Processing. 12, 453–461.

Ahmed, J., Memon, R. A., Waqas, M., et al., 2018. Selective sparse coding based coupled dictionary learning algorithm for single image super-resolution. in: Proc. of International Conference on Computing, Mathematics and Engineering Technologies, pp. 1-5.

Bevilacqua, M., Roumy, A., Guillemot, C., Morel, M.L.A., 2012. Low-complexity single-image super-resolution based on nonnegative neighbor embedding. in: Proc. of British Machine Vision Conference, pp. 1-10.

Cui, Z., Chang, H., Shan, S., Zhong, B., 2014. Deep network cascade for image super-resolution. in: Proc. of IEEE International Conference on Computer Vision, pp. 49–64.

Choi, J. H., Kim, J. H., et al., 2018. Deep Learning-based Image Super-Resolution Considering Quantitative and Perceptual Quality. in: Proc. of IEEE Conf. on Computer Vision and Pattern Recognition. 347-359

Dong, C., Loy, C.C., He, K., Tang, X., 2014. Learning a deep convolutional network for image super-resolution. in: Proc. of European Conference on Computer Vision, pp. 184–199.

Dong, C., Loy, C.C., He, K., et al., 2014. Image Super-Resolution Using Deep Convolutional Networks. IEEE Trans. IEEE Trans Pattern Anal Mach Intell. 38, 295–307.

Dong, C., Loy, C.C., and X, Tang., 2016. Accelerating the super resolution convolutional neural network. in: Proc. of European Conference on Computer Vision, pp. 391-407.

Farhadifard, F., Abar, E., Nazzal, M., et al., 2014. Single image super resolution based on sparse representation via directionally structured dictionaries. in: Proc. of Signal Processing and Communications Applications Conference, pp. 1718 – 1721.

Goodfellow, I. J., Pouget-Abadie, J., Mirza, M., et al., 2014. Generative Adversarial Nets. in: Proc. of International Conference on Neural Information Processing Systems, pp. 2672-2680

He, K., Zhang, X., Ren, S., et al., 2016. Deep Residual Learning for Image Recognition. in: Proc. of IEEE Conf. on Computer Vision and Pattern Recognition, pp. 770-778.

He, K., Zhang, X., Ren, S., Sun, J., 2015. Delving deep into rectifiers: Surpassing human-level performance on image-net classification. in: Proc. of IEEE International Conference on Computer Vision, pp. 1026–1034.

Kim, J., Lee, J.K., Lee, K.M., 2016. Accurate Image Super-Resolution Using Very Deep Convolutional Networks. in: Proc. of IEEE Conf. on Computer Vision and Pattern Recognition, pp. 1646-1654.

Lin-Yuan, Z., Cai-Xia, S. U., 2016. Image super-resolution via sparse representation. Computer Engineering and Design. IEEE Trans. IEEE Transactions on Image Processing.19, 2861 – 2873.

Ledig, C., Theis, L., Huszar, F., et al., 2017. Photo-Realistic Single Image Super-Resolution Using a Generative Adversarial Network. in: Proc. of IEEE International Conference on Computer Vision, pp. 4681-4690.

Lim, B., Son, S., Kim, H., et al., 2017. Enhanced Deep Residual Networks for Single Image Super-Resolution. in: Proc. of IEEE Conference on Computer Vision and Pattern Recognition Workshops, pp. 136-144.

Lai, W. S., Huang, J. B., Ahuja, N., et al., 2017. Fast and Accurate Image Super-Resolution with Deep Laplacian Pyramid Networks. IEEE Trans. Pattern Analysis and Machine Intelligence. 41, 2599 – 2613.

Martin, D., Fowlkes, C., Tal, D., et al., 2001. A Database of Human Segmented Natural Images and its Application to Evaluating Segmentation Algorithms and Measuring Ecological Statistics. in: Proc. of IEEE International Conference on Computer Vision, pp. 416–423.

Pang, J., Sun, W., Ren, J.S., 2017. Cascade Residual Learning: A Two-stage Convolutional Neural Network for Stereo Matching. in: Proc. of IEEE Conf. on Computer Vision and Pattern Recognition, pp. 887-895.

Peng, Y., 2019. Super-resolution Reconstruction Using Multiconnection Deep Residual Network Combined an Improved Loss Function for Single-frame Image. IEEE Trans. Multimedia Tools and Applications. 78, pp. 1–12.

Timofte, R., De, V., Gool, L. V., 2013. Anchored Neighborhood Regression for Fast Example-Based Super-Resolution. in: Proc. of IEEE International Conference on Computer Vision, pp. 1920–1927.

Timofte, R., Vincent, De., Smet, Luc. Van. Gool., 2014. A+: Adjusted Anchored Neighborhood Regression for Fast Super-Resolution. in: Proc. of IEEE Conf. on Computer Vision and Pattern Recognition, pp. 111–126.

Wang, Z., Bovik, A. C., Sheikh, H. R., et al., 2004. Image Quality Assessment: From Error Visibility to Structural Similarity. IEEE Trans. Image Processing. 13, 600-612.

Wang, Z., Liu, D., Yang, J., Han, W., Huang, T., 2015. Deeply improved sparse coding for image super-resolution. in: Proc. of IEEE International Conference on Computer Vision, pp.370–378.

Wang, X., Yu, K., Wu, S., et al., 2018. ESRGAN: Enhanced Super-Resolution Generative Adversarial Networks. in: Proc. of IEEE Conf. on Computer Vision and Pattern Recognition.

Wang, Z., Chen, J., Hoi, S. C. H., 2019. Deep Learning for Image Super-resolution: A Survey. IEEE Trans. Computer Vision and Pattern Recognition.

Yang, J., Wright, J., Huang, T. S., et al., 2008. Image super-resolution as sparse representation of raw image patches. in: Proc. of IEEE Conf. on Computer Vision and Pattern Recognition, pp. 2861–2873.

Yang, J., Wang, Z., Lin, Z., et al., 2012. Coupled Dictionary Training for Image Super-Resolution. IEEE Trans. Image Processing. 21, 3467 – 3478.

Yang, C.Y., Yang, M.H., 2013. Fast direct super-resolution by simple functions. in: Proc. of IEEE International Conference on Computer Vision, pp. 561–568.

Yi, C., Huang, J., 2015. Semismooth Newton Coordinate Descent Algorithm for Elastic-Net Penalized Huber Loss Regression and Quantile Regression. IEEE Trans. Statistics. 26, 547-557.

Yun, S., J, Choi., Y, Yoo., K, Yun., 2017. Action decision networks for visual tracking with deep reinforcement learning. in: Proc. of IEEE Conf. on Computer Vision and Pattern Recognition, pp. 2711-2720.

Yu, K., Dong, C., Lin, L., Loy, C.C., 2018. Crafting a toolchain for image restoration by deep reinforcement learning. in: Proc. of IEEE Conf. on Computer Vision and Pattern Recognition, pp. 2443-2452.

Yuan, Y., Liu, S., Zhang, J., et al., 2018. Unsupervised Image Super-Resolution using Cycle-in-Cycle Generative Adversarial Networks. in: Proc. of IEEE Conference on Computer Vision and Pattern Recognition Workshops, pp. 2443-2452.

Yang, X., Wu, W., Lu, L., et al., 2019. Multiple Regressions based Image Super-resolution. IEEE Trans. Multimedia Tools and Applications, pp. 1-17.

Zhang, G., Sun, H., Zheng, Y., et al., 2019. Optimal Discriminative Projection for Sparse Representation-based Classification via Bilevel Optimization. IEEE Trans. Circuits and Systems for Video Technology.

Zhang, G., Zheng, Y., Xia, G., 2019. Domain adaptive collaborative representation based classification. Multimedia Tools and Applications. 78, 175-196.

Zhang, G., Sun, H., Porikli, F., et al., 2017. Optimal couple projections for domain adaptive sparse representation-based classification. IEEE Trans. Image Processing. 26, 5922-5935.

Zeyde, Roman., M, Elad., M, Protter., 2010. On Single Image Scale-Up Using Sparse-Representations. Curves and Surfaces. in: Proc. of Curves and Surfaces - 7th International Conference, Avignon, France, June 24-30, 2010, Revised Selected Papers, pp. 711–730.

Zeiler, M. D., Fergus, R., 2014. Visualizing and Understanding Convolutional Networks. in: Proc. of European Conference on Computer Vision. pp. 818–833.

Zhao, P., Wang, W., Lu, Y., et al., 2018. Transfer robust sparse coding based on graph and joint distribution adaption for image representation. IEEE Trans. Knowledge-Based Systems. 147, 1-11.

Zhang, Y., Tian, Y., Kong, Y., et al., 2018. Residual Dense Network for Image Super-Resolution. in: Proc. of IEEE Conf. on Computer Vision and Pattern Recognition, pp. 2472-2481.

Zheng, Y., Wang, X., Zhang, G., et al., 2019. Multiple Kernel Coupled Projections for Domain Adaptive Dictionary Learning. IEEE Trans. Multimedia. 21, 2292-2304.

Zhang, W., Liu, Y., Dong, C., et al., 2019. RankSRGAN: Generative Adversarial Networks with Ranker for Image Super-Resolution. in: Proc. of IEEE International Conference on Computer Vision, pp. 3096-3105.